\PassOptionsToPackage{table}{xcolor}
\documentclass{article} 
\usepackage[OT1]{fontenc}
\usepackage{iclr2027_conference,times}

\usepackage{amsmath,amsfonts,bm}

\def\eqref#1{equation~\ref{#1}}

\def\1{\bm{1}}

\DeclareMathAlphabet{\mathsfit}{\encodingdefault}{\sfdefault}{m}{sl}
\SetMathAlphabet{\mathsfit}{bold}{\encodingdefault}{\sfdefault}{bx}{n}

\usepackage[hidelinks]{hyperref}
\usepackage{needspace,etoolbox}
\usepackage{url}
\usepackage[textwidth=5.5in,textheight=9in,centering]{geometry}
\usepackage{booktabs,tabularx,array,makecell,amssymb,amsmath,float,multirow,subcaption,graphicx,enumitem,wrapfig}
\usepackage[table]{xcolor}
\usepackage{capt-of}
\usepackage{caption}
\usepackage{marvosym}
\usepackage{tikz}
\usetikzlibrary{arrows.meta}
\renewcommand{\arraystretch}{1.18}

\newcolumntype{L}{>{\raggedright\arraybackslash}X}

\usepackage{tabularx}
\usepackage{array}
\newcolumntype{R}{>{\raggedleft\arraybackslash}X}

\definecolor{projectblue}{HTML}{2D658C}
\newcommand{\projectpageicon}{%
    \tikz[baseline=-0.6ex]{%
        \draw[projectblue,line width=0.35pt] (0,0) circle[radius=3.8pt];
        \draw[projectblue,line width=0.35pt] (0,0) ellipse[x radius=1.7pt,y radius=3.8pt];
        \draw[projectblue,line width=0.35pt] (0,0) ellipse[x radius=3.8pt,y radius=1.7pt];
        \draw[projectblue,line width=0.35pt] (-3.8pt,0) -- (3.8pt,0);
    }%
}

\newcommand{\papertitle}{Skeleton-Grounded Latent Action Model for Robot World Models}
\title{\papertitle}

\author{Qi Lv$^{1,2}$\qquad Jiayi Lin$^{1,3}$\qquad Yuxin Jiang$^{2}$\qquad Jianzong Wang$^{3}$\\
\textbf{Xiang Deng$^{1}$\qquad Mike Shou Zheng$^{2}$\textsuperscript{\Letter}} \\
$^{1}$HIT (Shenzhen)\quad $^{2}$Show Lab, NUS \quad $^{3}$Ping An Tech \\[4pt]
{\normalfont\small \projectpageicon\enspace Project page: {\color{magenta}\url{https://aopolin-lv.github.io/SkeLAM/}}}
}

\makeatletter
\patchcmd{\@maketitle}{\vskip 0.3in minus 0.1in}{\vskip 6pt}{}{%
    \PackageError{skelam}{Unable to adjust title-block spacing}{Check the conference template's maketitle definition.}}
\makeatother

\iclrfinalcopy
\begin{document}

\maketitle
\lhead{\papertitle}

\begin{abstract}
Robot world models simulate how environments evolve in response to robot actions. 
Latent action models enable learning dynamics from videos by encoding visual transitions as proxy actions. 
However, reconstruction-driven latent actions mix robot motion with object responses and other scene changes, complicating their correspondence with robot controls. 
We propose \textbf{SkeLAM}, a skeleton-guided latent action model for controllable robot world modeling. 
SkeLAM decomposes latent actions into a kinematic component grounded by robot skeleton supervision and a residual component retaining complementary visual dynamics. 
Cross-modal swap reconstruction encourages RGB-derived and skeleton-derived kinematic representations to be interchangeable during video reconstruction. 
The world model incorporates both components through separate conditioning pathways, while branch dropout reduces its dependence on future-derived residual information.
A lightweight adapter maps robot actions to the kinematic space, and subsequent world-model fine-tuning adapts generation to these predicted conditions. 
At inference, the model generates future observations from the current observation and robot actions. 
On DROID, SkeLAM reduces end-effector ADE by 19.5\% and LPIPS by 11.1\% compared with DreamDojo. 
Motion probing supports kinematic specialization, while qualitative results demonstrate zero-shot human-to-robot motion transfer in generated rollouts. 
Policy evaluation shows agreement between predicted success rates and those measured in real robot executions.
\end{abstract}

\section{Introduction}
Robot world models simulate how environments evolve in response to robot actions, supporting policy evaluation and model-based planning~\citep{yang2024unisim,wu2024ivideogpt}.
Accurate simulation requires modeling robot motion and resulting scene changes.
Latent action models encode visual transitions as proxy actions~\citep{bruce2024genie,ye2025lapa}, allowing world models to learn dynamics from large-scale videos without explicit action annotations~\citep{gao2026dreamdojo}.
These latent actions are learned through future-observation reconstruction, with the current observation providing visual context~\citep{gao2025adaworld,liang2026clam}.
For action-conditioned simulation, inferred latent actions serve as supervision for mapping robot controls into the learned latent action space.

Learning this mapping is challenging since reconstruction objectives encourage latent actions to encode robot motion together with object responses and other scene changes~\citep{nikulin2025distractors,zhang2025lamlearn}.
For example, similar end-effector motions may produce different object responses under different contact states~\citep{gu2026geniworld,xu2026kinema4d}, introducing scene-dependent variation beyond robot motion into the adaptation target.
These interaction outcomes are useful for world modeling, since accurate simulation requires predicting how robot motion affects the surrounding scene.
This motivates a dedicated kinematic component for action adaptation, together with a residual component retaining complementary transition information for dynamics learning.

To this end, we propose SkeLAM, a skeleton-guided latent action model for controllable robot world modeling.
Robot skeletons describe robot configuration and motion while abstracting scene appearance.
Related robot world models use skeleton sequences or kinematic renderings as direct conditioning signals for generation~\citep{wang2025vap,wu2026oscar,xu2026kinema4d}.
SkeLAM uses paired RGB and skeleton transitions to learn kinematic and residual components from RGB observations.
Skeleton supervision grounds the kinematic latent in robot motion, while the residual latent supplies complementary information for RGB reconstruction.
Together, the two components jointly support reconstruction, providing a dedicated target for action adaptation while retaining information for dynamics learning.
During representation learning, skeletons supervise the RGB latents, while the deployed interface uses RGB observations and robot actions.

We further introduce cross-modal swap reconstruction, which replaces the RGB kinematic latent with its paired skeleton counterpart while retaining the RGB residual.
Reconstructing the same transition under this substitution encourages compatible motion conditioning in the RGB decoder, extending cross-modal correspondence beyond alignment in latent space.
The world model learns from both components through separate conditioning pathways, while branch dropout prepares it for the zero-residual condition used at deployment.
A lightweight adapter maps robot actions to the kinematic space, and subsequent world-model fine-tuning adapts generation to these predicted conditions.
At inference, generation uses the current observation and robot actions.

To assess motion accuracy and visual fidelity, we evaluate SkeLAM on DROID~\citep{khazatsky2024droid}.
Compared with DreamDojo, SkeLAM reduces end-effector ADE by 19.5\% and LPIPS by 11.1\%.
Motion probing shows that the kinematic component approaches the performance of the concatenated representation with half the latent dimensions, while latent interventions and component ablations support its role in generation.
Qualitative results further demonstrate zero-shot human-to-robot motion transfer in generated rollouts.
Open-loop policy evaluation shows agreement between predicted success rates and those measured in real robot executions.

Our contributions are summarized as follows:
\begin{itemize}[leftmargin=10pt]
    \item We propose SkeLAM, a skeleton-guided latent action model that separates the conditioning roles of robot motion and complementary visual dynamics, providing a dedicated kinematic target for robot action adaptation.

    \item We introduce cross-modal swap reconstruction to encourage compatible RGB and skeleton motion conditioning during generation.
    Combined with branch dropout, this design supports robot action conditioning through the
    kinematic pathway learned from visual transitions.

    \item Experiments on DROID demonstrate gains in motion accuracy and visual fidelity, supported by representation analyses and component ablations.
    We further show zero-shot human-to-robot motion transfer and agreement between predicted and measured policy success rates.
\end{itemize}

\section{Related Work}
\label{sec:related_work}

\textbf{Action-conditioned robot world models.}
Robot world models support interactive simulation~\citep{yang2024unisim,wu2024ivideogpt,wang2025hma} and policy evaluation~\citep{zhu2025irasim,guo2026ctrlworld,liao2025genieenvisioner}.
Visual prediction supports foresight-guided policy learning~\citep{lv2025f1,cai2026internvla} and multi-view data augmentation~\citep{wang2026robovip}.
Related research addresses video procedure planning~\citep{ye2026osef}, temporal answer grounding~\citep{zhang2026difficulty}, and cloud--edge agent execution~\citep{huang2026agent}.
DALI-R~\citep{luo2026dalir} reranks actions with 3D point-cloud latent predictions and task-completion scores.
Control representations include action embeddings, skeleton sequences~\citep{wang2025vap,wu2026oscar}, and kinematic renderings~\citep{xu2026kinema4d}.
Recent models further incorporate spatial grounding~\citep{gu2026geniworld,yang2026lawm3d}.
Separating camera and object motion also supports camera-controlled video generation~\citep{cheng2026moca}.

\textbf{Latent action learning from videos.}
Latent action models encode visual transitions for world modeling~\citep{bruce2024genie,gao2025adaworld,gao2026dreamdojo,garrido2026latentworld} and robot policy learning~\citep{ye2025lapa,liang2026clam,yang2026como}.
Reconstruction-based learning is sensitive to distracting dynamics and ambiguous action correspondence~\citep{nikulin2025distractors,zhang2025lamlearn}.
Olaf-World~\citep{jiang2026olaf} aligns integrated latent actions with temporal differences in frozen video features across contexts.
LAOF~\citep{bu2026laof} uses agent optical flow as motion supervision to reduce sensitivity to distractors.
Shared RGB--depth features provide geometric supervision~\citep{govind2026unilact}, alongside advances in latent representations for policy learning~\citep{liu2026wala,shen2026ld4wam,xie2026geolam,liu2026gala}.
LARYBench~\citep{nie2026lary} evaluates representations using semantic action classification and physical action regression, independently of downstream policies.
A systematic study finds that proxy metrics support coarse model selection but poorly rank downstream policy performance~\citep{bu2026matters}.
SkeLAM grounds its kinematic component in robot skeletons while retaining complementary visual dynamics in a separate residual.

\textbf{Structured action representations.}
Quantized skills support autoregressive action generation~\citep{li2025star}, while flow-matching policies preserve latent-action structure and align latent and physical actions through interpolation-based sampling during decoder training~\citep{lyu2026lafp}.
Spatial and kinematic priors include voxel heatmaps~\citep{yang2026actionmap}, robot graphs with differentiable kinematics~\citep{lv2025kstar}, and trajectories defined by endpoint positions and velocities~\citep{lv2026hermite}.
Joint image, LiDAR, and action modeling supports autonomous-driving trajectory generation~\citep{cheng2026mojito}.
SkeleWAM~\citep{sheng2026skelewam} jointly learns action and future-state prediction over sparse 3D robot--object skeletons.
Skel-WAM~\citep{cai2026skelwam} cotrains visual and hand-skeleton dynamics on human and robot videos, with a robot-trained action expert.
SkeLAM uses skeleton supervision to ground video-derived kinematic latents for action-conditioned video prediction.

\begin{figure}[t]
    \centering
    \includegraphics[width=\linewidth]{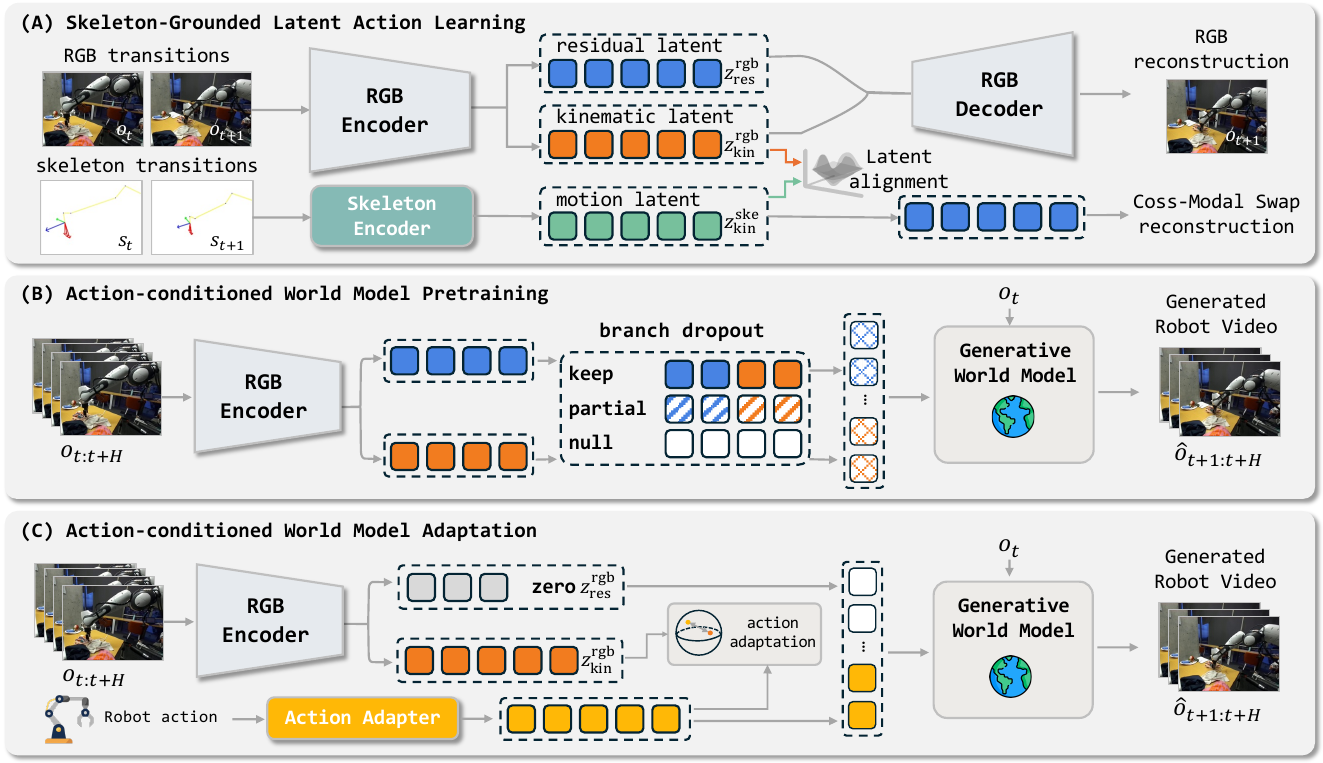}
    \caption{
        \textbf{Overview of SkeLAM.}
        (A) Skeleton supervision and cross-modal swap reconstruction structure RGB transition latents into kinematic and residual components.
        (B) Separate projections condition the world model: keep retains both branches, partial drops one, and null drops both.
        (C) An action adapter predicts kinematic latents, with the null residual at inference.
    }
    \label{fig:method_overview}
\end{figure}

\section{Methodology}
\label{sec:methodology}

SkeLAM learns structured latent actions for video-based world modeling and subsequent adaptation to robot controls.
As illustrated in Figure~\ref{fig:method_overview}, skeleton supervision grounds a kinematic component in robot motion, while a residual component retains complementary visual dynamics.
The world model learns from both components, with branch dropout preparing it for action-conditioned inference.
An action adapter is then trained to predict kinematic latents from robot actions, followed by world-model fine-tuning using these predictions.

\textbf{Preliminary.}
Let $o_t$ and $a_t$ denote the visual observation and robot action at time $t$.
An action-conditioned world model represents the distribution of $H$ future observations:
\begin{equation}
    p\!\left(
        o_{t+1:t+H}
        \mid o_t,a_{t:t+H-1}
    \right).
    \label{eq:world_model_preliminary}
\end{equation}
To learn dynamics from videos, latent action models infer proxy actions $z_t$ from visual transitions $(o_t,o_{t+1})$.
These representations are learned by reconstructing $o_{t+1}$ from $o_t$ and $z_t$, and subsequently provide conditioning signals for world-model training.

\subsection{Skeleton-Grounded Latent Actions}
\label{sec:structured_latents}

\textbf{Latent decomposition.} To learn a kinematic representation for robot action adaptation while retaining complementary information for RGB reconstruction, we encode each RGB transition into kinematic latents $z_{\mathrm{kin},t}^{\mathrm{rgb}}$ and residual latents $z_{\mathrm{res},t}^{\mathrm{rgb}}$.
Further, to supervise the kinematic latent, we use skeleton images depicting the corresponding robot joint configurations.

Given paired RGB observations $(o_t,o_{t+1})$ and their corresponding skeleton images $(s_t,s_{t+1})$, Transformer encoders parameterize Gaussian posteriors for the RGB kinematic and residual latents, and the skeleton motion latent:
\begin{equation}
    \begin{aligned}
        \bigl(z_{\mathrm{kin},t}^{\mathrm{rgb}},
              z_{\mathrm{res},t}^{\mathrm{rgb}}\bigr)
        &\sim q_{\phi}^{\mathrm{rgb}}
        (z_{\mathrm{kin}},z_{\mathrm{res}}\mid o_t,o_{t+1}), \\
        z_{\mathrm{kin},t}^{\mathrm{ske}}
        &\sim q_{\eta}^{\mathrm{ske}}
        (z_{\mathrm{kin}}\mid s_t,s_{t+1}).
    \end{aligned}
    \label{eq:structured_encoding}
\end{equation}
where $z_{\mathrm{kin},t}^{\mathrm{rgb}},z_{\mathrm{kin},t}^{\mathrm{ske}} \in\mathbb{R}^{d_{\mathrm{kin}}}$ and $z_{\mathrm{res},t}^{\mathrm{rgb}} \in\mathbb{R}^{d_{\mathrm{res}}}$.
The $d_{\mathrm{kin}}$ and $d_{\mathrm{res}}$ denote the dimensions of the kinematic and residual latents, respectively.
We initialize the RGB and skeleton branches through conditional reconstruction:
\begin{equation}
    \begin{aligned}
        \hat{o}_{t+1}
        &= D_{\mathrm{rgb}}\!\left(
            o_t,z_{\mathrm{kin},t}^{\mathrm{rgb}},
            z_{\mathrm{res},t}^{\mathrm{rgb}}\right), \\
        \hat{s}_{t+1}
        &= D_{\mathrm{ske}}\!\left(
            s_t,z_{\mathrm{kin},t}^{\mathrm{ske}}\right).
    \end{aligned}
    \label{eq:rgb_reconstruction}
\end{equation}
Here, skeleton reconstruction establishes the motion reference used for subsequent supervision of the RGB kinematic representation.

\begin{wrapfigure}{r}{0.54\columnwidth}
    \vspace{-2.1em}
    \centering
    \includegraphics[width=\linewidth]{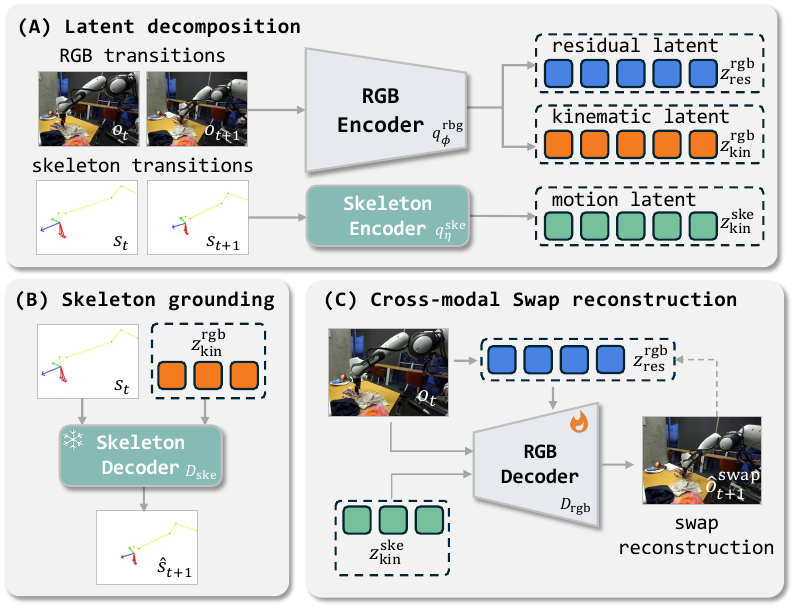}
    \caption{
        \textbf{Skeleton-grounded latent action learning.}
        (A) RGB and skeleton transition encoding.
        (B) Grounding through a frozen skeleton decoder.
        (C) Swap reconstruction using the paired skeleton mean and the same RGB residual sample.
    }
    \label{fig:latent_routing}
    \vspace{-3.5em}
\end{wrapfigure}

\textbf{Skeleton grounding.}
To ground the RGB kinematic representation in robot motion, we freeze the trained skeleton encoder and decoder.
We reconstruct the next skeleton from the RGB kinematic posterior mean through the frozen decoder and align it with the paired skeleton motion representation.
Let $u_t=\operatorname{norm}(\mu_{\mathrm{kin},t}^{\mathrm{rgb}})$ and $v_t=\operatorname{sg}[\operatorname{norm}(\mu_{\mathrm{kin},t}^{\mathrm{ske}})]$, where $\mu$ is the posterior mean, $\operatorname{norm}$ denotes $\ell_2$ normalization, and $\operatorname{sg}$ denotes stop-gradient.
The grounding and alignment objectives are:
\begingroup
\small
\begin{equation}
    \begin{aligned}
        \mathcal L_{\mathrm{ground}}
         &=\mathbb E\!\left[\ell_{\mathrm{ske}}\!\left(
               D_{\mathrm{ske}}(s_t,\mu^{\mathrm{rgb}}_{\mathrm{kin},t}),s_{t+1}\right)\right],\\
        \mathcal L_{\mathrm{align}}
         &=\mathbb E\!\left[1-u_t^{\mathsf T}v_t
            +\frac{\|u_t-v_t\|_2^2}{d_{\mathrm{kin}}}\right].
    \end{aligned}
    \label{eq:skeleton_grounding}
\end{equation}
\endgroup
Here, $\ell_{\mathrm{ske}}$ is the skeleton reconstruction loss detailed in Appendix~\ref{app:stage1_training_details}.

\textbf{Cross-modal swap reconstruction.}
To extend motion supervision from the skeleton branch to the RGB decoder, we introduce cross-modal swap reconstruction.
We replace the RGB kinematic input with the posterior mean of the paired skeleton motion latent, while retaining the residual sample used in RGB reconstruction:
\begin{equation}
    \begin{aligned}
        \hat{o}_{t+1}^{\mathrm{swap}}
        &= D_{\mathrm{rgb}}\!\left(
            o_t,
            \operatorname{sg}\!\left[
                \mu_{\mathrm{kin},t}^{\mathrm{ske}}\right],
            z_{\mathrm{res},t}^{\mathrm{rgb}}
        \right), \\
        \mathcal{L}_{\mathrm{swap}}
        &= \mathbb{E}\!\left[
            \operatorname{MAE}\!\left(
                \hat{o}_{t+1}^{\mathrm{swap}},o_{t+1}
            \right)
        \right].
    \end{aligned}
    \label{eq:cross_modal_swap}
\end{equation}
The RGB and swapped reconstructions share the decoder, current observation, and target, encouraging compatible motion conditioning across modalities.
Gradients propagate to $\phi$ through the residual latent, while the skeleton branch remains frozen.

\textbf{Latent-action objective.}
The RGB objective combines reconstruction MSE with KL regularization of the kinematic and residual posteriors toward standard Gaussian priors:
\begin{equation}
    \begin{aligned}
        \mathcal{L}_{\mathrm{rgb}}
        ={}&
        \mathbb{E}_{\mathcal{D},q_{\phi}^{\mathrm{rgb}}}\!\left[
            \operatorname{MSE}(\hat{o}_{t+1},o_{t+1})
        \right] \\
        &+
        \beta\mathbb{E}_{\mathcal{D}}\!\left[
            D_{\mathrm{KL}}\!\left(
                q_{\phi,\mathrm{kin}}^{\mathrm{rgb}}
                \,\Vert\,\mathcal{N}(0,I)
            \right)
            +
            D_{\mathrm{KL}}\!\left(
                q_{\phi,\mathrm{res}}^{\mathrm{rgb}}
                \,\Vert\,\mathcal{N}(0,I)
            \right)
        \right].
    \end{aligned}
    \label{eq:rgb_variational_objective}
\end{equation}
Here $\mathcal{D}$ denotes the paired training data and $\beta$ controls the KL regularization strength.
The component posteriors $q_{\phi,\mathrm{kin}}^{\mathrm{rgb}}$ and $q_{\phi,\mathrm{res}}^{\mathrm{rgb}}$ are the marginals of $q_{\phi}^{\mathrm{rgb}}(\cdot\mid o_t,o_{t+1})$.

With the skeleton branch frozen, we optimize the RGB encoder and decoder using:
\begin{equation}
    \begin{aligned}
        \mathcal{L}_{\mathrm{latent}}
        ={}&
        \mathcal{L}_{\mathrm{rgb}}
        +\lambda_{\mathrm{ground}}
         \mathcal{L}_{\mathrm{ground}}
        +\lambda_{\mathrm{align}}
         \mathcal{L}_{\mathrm{align}} \\
        &+
        \lambda_{\mathrm{swap}}
         \mathcal{L}_{\mathrm{swap}}
        +\lambda_{\mathrm{var}}
         \mathcal{L}_{\mathrm{var}}.
    \end{aligned}
    \label{eq:latent_training_objective}
\end{equation}
The regularizer $\mathcal{L}_{\mathrm{var}}$ penalizes low per-dimension standard deviations and cross-covariance between the RGB latent components.
Grounding and latent alignment are introduced after conditional reconstruction initialization, followed by swap reconstruction.
Initialization losses, regularizer definitions, loss weights, and the optimization schedule are provided in Appendix~\ref{app:stage1_training_details}.

\subsection{Action-Conditioned World Modeling}
\label{sec:action_conditioned_world_model}

\textbf{Structured world-model conditioning.}
To preserve a kinematic conditioning pathway for robot action adaptation, we introduce separate projections for the kinematic and residual components.
Given a training clip $o_{t:t+H}$, the frozen RGB encoder extracts $z_{\mathrm{kin},i}^{\mathrm{rgb}}$ and $z_{\mathrm{res},i}^{\mathrm{rgb}}$ from each transition $(o_i,o_{i+1})$, where $i\in\{t,\ldots,t+H-1\}$.
We use separate MLP projections $P_b$ with branch-presence masks $m_b\in\{0,1\}$, leaving latent standardization implicit in the notation (Appendix~\ref{app:latent_cache}):
\begin{equation}
    \begin{aligned}
        \tilde z_{b,i}
        &=m_b z^{\mathrm{rgb}}_{b,i},\qquad b\in\{\mathrm{kin},\mathrm{res}\},\\
        h_i
        &= P_{\mathrm{kin}}\!\left(
            \tilde z_{\mathrm{kin},i};m_{\mathrm{kin}}\right)
        + P_{\mathrm{res}}\!\left(
            \tilde z_{\mathrm{res},i};m_{\mathrm{res}}\right),
    \end{aligned}
    \label{eq:structured_conditioning}
\end{equation}
where a dropped branch has a zero latent and a zero presence mask.
The world model $p_\theta(o_{t+1:t+H}\mid o_t,h_{t:t+H-1})$ learns from video-derived latents through full-backbone fine-tuning after projector warmup.
Robot-action latents reuse the kinematic projection.

\textbf{Branch dropout and world-model training.}
Residual latents use future observations unavailable at deployment.
To retain complementary information while preparing for missing conditions, $\mathcal R$ samples one pair of branch masks per clip:
\begin{equation}
    (m_{\mathrm{kin}},m_{\mathrm{res}})=
    \begin{cases}
        (1,1), & \text{full (keep)},\\
        (0,1)\text{ or }(1,0),
            & \text{drop\_kin or drop\_res (partial)},\\
        (0,0), & \text{drop\_both (null)}.
    \end{cases}
    \label{eq:residual_dropout}
\end{equation}
These four whole-branch modes share masks across all $H$ transitions; their step-dependent probabilities are given in Appendix~\ref{app:stage2_details}.
Partial means dropping one entire branch, not individual coordinates; null drops both branches using fixed zeros in standardized latent space.
The supervision target remains the complete future clip $o_{t+1:t+H}$ in every mode.
The drop\_res mode trains prediction from $o_t$ and kinematic latents, matching action-conditioned inference.

Let $\rho$ collect the learnable conditioning parameters.
The world-model generation objective is
\begin{equation}
    \mathcal{L}_{\mathrm{wm}}(\theta,\rho)
    =
    \mathbb{E}_{
        \substack{
            \mathcal{D}_{\mathrm{video}},q_{\phi}^{\mathrm{rgb}},
            \mathcal{R},\xi
        }
    }\!\left[
        \ell_{\mathrm{gen}}\!\left(
            \theta;
            o_{t+1:t+H},o_t,h_{t:t+H-1},\xi
        \right)
    \right],
    \label{eq:world_model_objective}
\end{equation}
where $\mathcal{D}_{\mathrm{video}}$ contains training clips, $\ell_{\mathrm{gen}}$ is the generation loss, and $\xi$ collects its stochastic variables.
The RGB encoder remains frozen.

\textbf{Robot action adaptation and inference.}
To condition the world model on robot controls, we train a lightweight MLP adapter $A_{\psi}:\mathcal{A}\rightarrow\mathbb{R}^{d_{\mathrm{kin}}}$, where $\mathcal{A}$ denotes the robot action space.
For each transition $(o_i,a_i,o_{i+1})$, the frozen RGB encoder provides the kinematic posterior mean $\mu_{\mathrm{kin},i}^{\mathrm{rgb}}$ as the regression target:
\begin{equation}
    \mathcal{L}_{\mathrm{adapt}}(\psi)
    =
    \mathbb{E}_{\mathcal{D}_{\mathrm{robot}}}\!\left[
        d_{\mathrm{adapt}}\!\left(
            A_{\psi}(a_i),
            \operatorname{sg}\!\left[
                \mu_{\mathrm{kin},i}^{\mathrm{rgb}}
            \right]
        \right)
    \right],
    \label{eq:action_adaptation_objective}
\end{equation}
where $\mathcal{D}_{\mathrm{robot}}$ denotes the robot training data and $d_{\mathrm{adapt}}$ is the latent regression criterion.
Restricting regression to the kinematic component focuses adaptation on the motion representation learned through skeleton supervision.
Only $\psi$ is optimized, with the RGB encoder and $(\theta,\rho)$ held fixed.

We then fine-tune the full world-model backbone using predicted kinematic latents and a curriculum residual--mask pair $(\tilde z_{\mathrm{res},i}^{(s)},m_{\mathrm{res}}^{(s)})$ (Appendix~\ref{app:stage3_details}):
\begin{equation}
    \hat{h}_{i}^{\mathrm{act}}
    =
    P_{\mathrm{kin}}\!\left(A_{\psi}(a_i);1\right)
    +
    P_{\mathrm{res}}\!\left(\tilde z_{\mathrm{res},i}^{(s)};m_{\mathrm{res}}^{(s)}\right).
    \label{eq:action_derived_conditioning}
\end{equation}
The action-conditioned generation objective is
\begin{equation}
    \mathcal{L}_{\mathrm{wm}}
    =
    \mathbb{E}_{\mathcal{D}_{\mathrm{robot}},\xi}
    \!\left[
        \ell_{\mathrm{gen}}\!\left(
            \theta;
            o_{t+1:t+H},o_t,
            \hat{h}_{t:t+H-1}^{\mathrm{act}},\xi
        \right)
    \right].
    \label{eq:action_conditioned_training}
\end{equation}

At inference, $\tilde z_{\mathrm{res},i}=0$ and $m_{\mathrm{res}}=0$, with no future teacher residual; the current observation and actions specify the rollout:
\begin{equation}
    \hat{o}_{t+1:t+H}
    \sim
    p_{\theta}\!\left(
        \,\cdot\mid o_t,
        \hat{h}_{t:t+H-1}^{\mathrm{act}}
    \right).
    \label{eq:action_conditioning}
\end{equation}

\section{Experiment}
We evaluate the effect of the kinematic--residual decomposition on action-conditioned world modeling, and examine motion transfer and rollout evaluation as downstream applications.
Specifically, we investigate the following questions:
\begin{itemize}[leftmargin=1.5em]

    \item \textbf{RQ1 (World modeling):}
    Does adapting robot actions to a dedicated kinematic component improve motion accuracy while preserving visual fidelity? (Section~\ref{sec:skelam_main_results})

    \item \textbf{RQ2 (Motion representation):}
    Do skeleton-grounded kinematic latents support more accurate linear decoding of robot motion than unstructured latent actions? (Section~\ref{sec:motion_representation})

    \item \textbf{RQ3 (Motion transfer):}
    Can a kinematic representation learned from robot interactions transfer human motion to generated robot rollouts in a zero-shot setting? (Section~\ref{sec:skelam_human_transfer})

    \item \textbf{RQ4 (Policy evaluation):}
    How closely do open-loop world-model rollouts reproduce policy rankings and success rates measured in real robot executions? (Section~\ref{sec:skelam_policy_evaluation})
\end{itemize}

\subsection{Experimental Setup}
\textbf{Data.}
DROID~\citep{khazatsky2024droid} pairs visual observations with robot actions, allowing us to assess whether predicted dynamics follow robot controls.
We train the latent action model and world model on DROID, incorporating AgiBot~\citep{bu2025agibot} to broaden the range of robot interactions observed during training.
Evaluation uses 400 DROID clips, each containing one conditioning frame and 12 future frames at $480\times640$ resolution. 
More details are provided in Appendix~\ref{app:skelam_video_data}.

\textbf{Evaluation Metrics.} To quantify trajectory, direction, and magnitude agreement, we report 2D end-effector average displacement error (ADE, px), direction macro-F1 ($\mathrm{F1}_{\mathrm{dir}}$), normalized magnitude MAE ($L_1^{\mathrm{mag}}$), and magnitude Pearson correlation ($r_{\mathrm{mag}}$).
PSNR, SSIM, and LPIPS~\citep{zhang2018perceptual} assess pixel, structural, and perceptual fidelity, respectively, with tLPIPS reported as a temporal diagnostic.
More details are provided in Appendix~\ref{app:skelam_video_metrics}.

\textbf{Implementation Details.}
We implement the latent action model using Transformer VAEs with $d_{\mathrm{kin}}=d_{\mathrm{res}}=32$ and use Cosmos-Predict2.5-2B as the world-model backbone.
We train with AdamW, using stage-specific learning rates and global batch sizes.
SkeLAM training uses NVIDIA A100 GPUs and requires approximately 2 days for one complete run.
Architectural specifications, loss weights, training schedules, and inference settings are provided in Appendix~\ref{app:skelam_implementation}.

\textbf{Baselines.}
We compare SkeLAM with latent-action world models DreamDojo~\citep{gao2026dreamdojo} and CD-LAM~\citep{wei2026cdlam}.
Alternative conditioning strategies include direct robot actions in IRASim~\citep{zhu2025irasim}, Ctrl-World~\citep{guo2026ctrlworld}, Cosmos-Predict2.5~\citep{ali2025world}, Cosmos3~\citep{agarwal2026cosmos}, and GE-Sim~\citep{liao2025genieenvisioner}, and geometric guidance in EnerVerse-AC~\citep{jiang2025enerverseac} and OSCAR~\citep{wu2026oscar}.
TesserAct~\citep{zhen2025tesseract}, and GE-Base~\citep{liao2025genieenvisioner} serve as text-conditioned references for visual fidelity, with evaluation details provided in Appendix~\ref{app:skelam_protocol}.

\subsection{Action-Conditioned World Modeling}
\label{sec:skelam_main_results}

\textbf{Quantitative results.}
To assess how accurately SkeLAM simulates the consequences of robot actions, we evaluate motion accuracy and visual fidelity on DROID.
Table~\ref{tab:main_results} shows that SkeLAM ranks first on seven of eight metrics among the reported action-conditioned models, including all four visual fidelity metrics.
Compared with DreamDojo and CD-LAM, SkeLAM reduces ADE by 19.5\% and 23.7\%, respectively, alongside LPIPS reductions of 11.1\% and 17.4\%.
These improvements indicate closer agreement with both the target robot trajectories and visual observations.
Ctrl-World and OSCAR achieve higher direction F1, while SkeLAM attains lower trajectory error and stronger visual fidelity.
In particular, compared with Ctrl-World, SkeLAM reduces ADE from 42.96 to 41.15 px and LPIPS from 0.4029 to 0.1310.
The combined improvements in trajectory accuracy and visual fidelity support the generation of robot motion together with the associated scene dynamics.

\begin{table}[h]
    \centering
    \caption{
        \textbf{World modeling results.}
        Control denotes the inference-time conditioning interface.
        Text-conditioned models serve as visual references.
        \textbf{Bold} and \underline{underlined} values indicate the best and second-best action-conditioned results.
        $^\dagger$ denotes large-scale interaction pretraining.
    }
    \vspace{-3mm}
    \label{tab:main_results}

    \fontsize{8.5}{10.2}\selectfont
    \setlength{\tabcolsep}{2.5pt}
    \renewcommand{\arraystretch}{1.08}
    \begin{tabularx}{\linewidth}{@{}ll *{8}{R}@{}}
        \toprule[1.02pt]
        & & \multicolumn{4}{c}{Visual fidelity}
        & \multicolumn{4}{c}{Motion accuracy} \\
        \cmidrule(lr){3-6}\cmidrule(lr){7-10}
        Method & Control 
          & PSNR$\uparrow$ & SSIM$\uparrow$ & LPIPS$\downarrow$ & MSE$\downarrow$
          & ADE$\downarrow$ & $\mathrm{F1}_{\mathrm{dir}}\uparrow$ & $L_1^{\mathrm{mag}}\downarrow$ & $r_{\mathrm{mag}}\uparrow$ \\
        \midrule
        \rowcolor{gray!12}
        \multicolumn{10}{l}{\textit{Text-conditioned references}} \\
        TesserAct & Text
            & 15.08 & 0.7400 & 0.2561 & 0.0349 
            & 83.76 & 0.2950 & 0.0797 & 0.0823 \\
        GE-Base & Text
            & 12.51 & 0.4530 & 0.5864 & 0.0596 
            & 55.30 & 0.3006 & 0.0956 & 0.0504 \\
        \midrule
        \rowcolor{blue!6}
        \multicolumn{10}{l}{\textit{Action- and geometry-conditioned models}} \\
        IRASim & Action emb.
            & 11.55 & 0.4137 & 0.5993 & 0.0738 
            & 67.11 & 0.2616 & 0.0843 & 0.1193 \\
        Ctrl-World & Action emb.
            & 16.34 & 0.5910 & 0.4029 & 0.0255 
            & \underline{42.96} & \textbf{0.5888} & \underline{0.0511} & \underline{0.5154} \\
        Cosmos-Predict2.5 & Action emb.
            & 11.51 & 0.4209 & 0.6018 & 0.0758 
            & 60.76 & 0.2375 & 0.0839 & 0.1049 \\
        Cosmos3 & Action emb.
            & 16.98 & 0.7630 & 0.2186 & 0.0270 
            & 67.23 & 0.3278 & 0.0725 & 0.2518 \\
        EnerVerse-AC & Action map
            & 14.13 & 0.5157 & 0.4139 & 0.0415 
            & 59.63 & 0.2433 & 0.0982 & -0.0050 \\
        GE-Sim & Action map
            & 12.53 & 0.4550 & 0.4950 & 0.0593 
            & 55.57 & 0.2799 & 0.0941 & 0.0091 \\
        OSCAR & Skeleton
            & 12.57 & 0.4471 & 0.4788 & 0.0587 
            & 43.74 & \underline{0.5810} & 0.0530 & 0.5029 \\
        \midrule
        \rowcolor{teal!8}
        \multicolumn{10}{l}{\textit{Models with learned latent actions}} \\
        DreamDojo$^{\dagger}$ & $a$
            & \underline{18.91} & \underline{0.8057} & \underline{0.1473} & \underline{0.0168} 
            & 51.11 & 0.4989 & 0.0667 & 0.3335 \\
        CD-LAM$^{\dagger}$ & $a\to z$
            & 18.34 & 0.7910 & 0.1586 & 0.0181 
            & 53.90 & 0.4903 & 0.0690 & 0.3179 \\
        \textbf{SkeLAM (ours)} & $a\to z_{\mathrm{kin}}$
            & \textbf{19.35} & \textbf{0.8340} & \textbf{0.1310} & \textbf{0.0151}
            & \textbf{41.15} & 0.5465 & \textbf{0.0510} & \textbf{0.5277} \\
        \bottomrule[1.02pt]
    \end{tabularx}
    \vspace{-2mm}
\end{table}

\begin{figure}[h]
    \centering
    \vspace{-1mm}
    \includegraphics[width=0.93\linewidth]{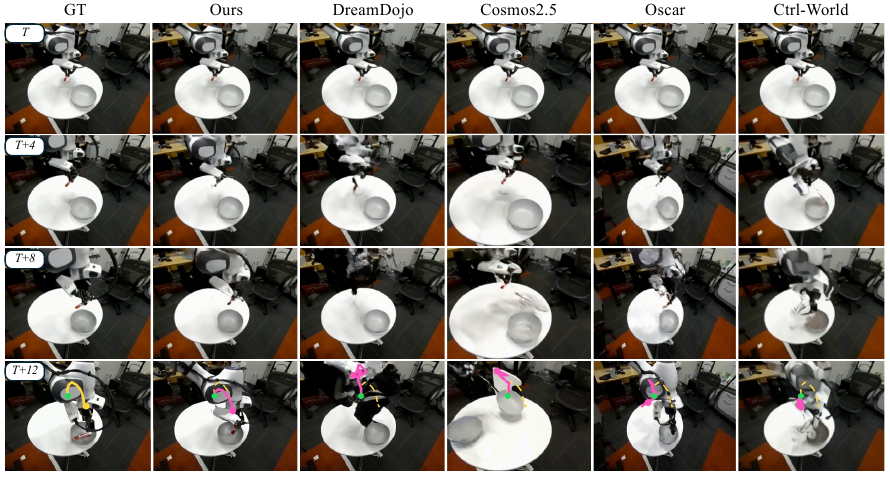}
    \caption{
        \textbf{Qualitative comparison of action-conditioned world modeling.}
        Ground-truth and generated observations are compared at selected time steps, highlighting differences in end-effector motion, manipulated-object geometry, and scene appearance.
    }
    \label{fig:qualitative_results}
    \vspace{-5mm}
\end{figure}

\textbf{Qualitative results.} 
To evaluate whether generated rollouts reproduce robot motion while preserving scene structure, we compare predictions with ground truth in Figure~\ref{fig:qualitative_results}.
The overlaid trajectories reveal deviations from the reference motion in DreamDojo and OSCAR, accompanied by visible arm deformation in later frames.
Ctrl-World produces a shorter trajectory and local gripper distortions, while Cosmos-Predict2.5 alters the apparent table scale and bowl position.
SkeLAM more closely follows the curved reference trajectory while retaining the tabletop layout, although local gripper errors remain.
This example highlights the importance of assessing motion correspondence alongside scene consistency.
Two additional comparisons appear in Appendix~\ref{app:qualitative_comparisons}.

\textbf{Ablation studies.}
To assess how each component supports action-conditioned generation, we compare variants in Table~\ref{tab:ablations}.
\textbf{(i) Skeleton supervision} grounds the RGB kinematic latent in robot motion; removing it increases ADE from 41.15 to 64.59 and reduces magnitude correlation from 0.5277 to 0.2009, supporting geometric supervision for learning motion representations suitable for action conditioning.
\textbf{(ii) Cross-modal swap} couples skeleton supervision to RGB generation beyond latent alignment; removing it increases ADE to 66.74 and reduces magnitude correlation to 0.1897, while LPIPS rises from 0.1310 to 0.1436, supporting reconstruction-level RGB--skeleton alignment for downstream action conditioning.
\textbf{(iii) Residual latent} retains complementary scene dynamics; zeroing it from world-model pretraining maintains similar visual fidelity (0.1305 vs.\ 0.1310 LPIPS) while ADE increases to 43.37 and magnitude correlation decreases to 0.4737, suggesting that complementary visual information benefits motion conditioning.
\textbf{(iv) Action adaptation} supervises the action-to-kinematic mapping; omitting adapter regression increases ADE to 65.84, supporting explicit supervision of the mapping from robot actions to the learned kinematic representation.
Further details and adaptation-budget analysis are provided in Appendix~\ref{app:additional_experiments}.

\begin{table}[t]
    \centering
    \caption{
        \textbf{Component ablations.} 
        Each variant removes one component from the full model under the same training budget. 
        \textbf{Bold} and \underline{underlined} values indicate the best and second-best results.
    }
    \label{tab:ablations}

    \small
    \setlength{\tabcolsep}{3pt}
    \renewcommand{\arraystretch}{1.}
    \begin{tabularx}{\linewidth}{
        @{}l*{8}{>{\raggedleft\arraybackslash}X}@{}
    }
        \toprule[1.02pt]
        & \multicolumn{4}{c}{Visual fidelity}
        & \multicolumn{4}{c}{Motion accuracy} \\
        \cmidrule(lr){2-5}\cmidrule(lr){6-9}
        Variant
        & PSNR$\uparrow$ & SSIM$\uparrow$
        & LPIPS$\downarrow$ & MSE$\downarrow$
        & ADE$\downarrow$
        & $\mathrm{F1}_{\mathrm{dir}}\uparrow$
        & $L_1^{\mathrm{mag}}\downarrow$
        & $r_{\mathrm{mag}}\uparrow$ \\
        \midrule
        \rowcolor{orange!9}
        \textbf{Full SkeLAM}
            & \underline{19.35} & \underline{0.8340}
            & \underline{0.1310} & \underline{0.0151}
            & \textbf{41.15} & \textbf{0.5465}
            & \textbf{0.0510} & \textbf{0.5277} \\
        \midrule
        w/o skeleton supervision
            & 18.18 & 0.8206
            & 0.1538 & 0.0188
            & 64.59 & 0.3268 & 0.0723 & 0.2009 \\
        w/o cross-modal swap
            & 19.30 & 0.8330 & 0.1436 & 0.0192
            & 66.74 & 0.3132 & 0.0752 & 0.1897 \\
        w/o residual latent
            & \textbf{19.38} & \textbf{0.8348}
            & \textbf{0.1305} & \textbf{0.0150}
            & \underline{43.37} & \underline{0.5263}
            & \underline{0.0547} & \underline{0.4737} \\
        w/o action adaptation
            & 18.14 & 0.8196
            & 0.1545 & 0.0189
            & 65.84 & 0.3259 & 0.0739 & 0.1603 \\
        \bottomrule[1.02pt]
    \end{tabularx}
    \vspace{4pt}
\end{table}

\subsection{Motion Representation Analysis}
\label{sec:motion_representation}

\begin{figure}[h]
    \centering
    \begin{minipage}[t]{0.44\linewidth}
        \vspace{0pt}
        \centering
        \captionsetup{font=small,skip=5pt,
            justification=raggedright,singlelinecheck=false}
        \captionof{table}{
            \textbf{Motion probing.}
            Linear probes predict EEF displacement (MAE, $R^2$) and motion magnitude (macro-F1) from frozen representations.
            Latents are 32D except the 64D concatenation.
            $^\dagger$: privileged skeleton input.
        }
        \label{tab:droid_motion_probing}
        {\fontsize{8.5}{10.2}\selectfont
        \setlength{\tabcolsep}{2.5pt}
        \renewcommand{\arraystretch}{1.07}
        \begin{tabularx}{\linewidth}{@{}>{\raggedright\arraybackslash}Xrrr@{}}
            \toprule[1.05pt]
            Representation
            & $\mathrm{MAE}\!\downarrow$
            & $R^2_{\mathrm{EEF}}\!\uparrow$
            & $\mathrm{F1}_{\mathrm{mag}}\!\uparrow$ \\
            \midrule
            DreamDojo & 0.01620 & 0.0078 & 0.7183 \\
            CD-LAM & 0.01615 & 0.0079 & 0.7006 \\
            \midrule
            \rowcolor{orange!9}
            \textbf{SkeLAM} $z_{\mathrm{kin}}$
                & 0.00599 & 0.3200 & 0.7467 \\
            SkeLAM $z_{\mathrm{res}}$
                & 0.00678 & 0.1879 & 0.5652 \\
            $[z_{\mathrm{kin}};z_{\mathrm{res}}]$
                & 0.00597 & 0.3277 & 0.7482 \\
            \midrule
            Skeleton teacher$^{\dagger}$
                & 0.00552 & 0.4312 & 0.7233 \\
            \bottomrule[1.05pt]
        \end{tabularx}\par}
    \end{minipage}%
    \hspace{0.03\linewidth}%
    \begin{minipage}[t]{0.53\linewidth}
        \vspace{0pt}
        \centering
        \includegraphics[width=\linewidth]{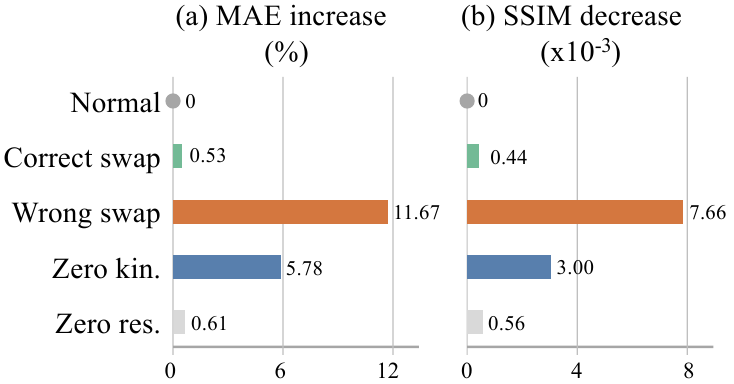}
        \captionsetup{font=small,skip=5pt,
            justification=raggedright,singlelinecheck=false}
        \captionof{figure}{
            \textbf{Latent action interventions.}
            Reconstruction MAE increase and SSIM decrease relative to Normal.
            Correct and wrong swaps use paired and mismatched skeleton kinematic means.
        }
        \label{fig:droid_kinematic_swap}
    \end{minipage}
\end{figure}

\textbf{Motion decodability.}
To examine how robot motion is encoded, we train linear probes on frozen latent representations.
As shown in Table~\ref{tab:droid_motion_probing}, the kinematic component achieves an EEF MAE of $5.99\times10^{-3}$ and a magnitude macro-F1 of 0.7467, improving on both latent-action baselines.
Within SkeLAM, it also outperforms the residual component on all three metrics.
Concatenating the residual increases EEF $R^2$ from 0.3200 to 0.3277 and magnitude macro-F1 from 0.7467 to 0.7482.
The kinematic component thus approaches the probing performance of the full representation using half the latent dimensions, supporting its specialization for displacement and magnitude encoding.

\textbf{Cross-modal compatibility.}
To assess cross-modal motion conditioning, we intervene on the latent inputs to the RGB decoder.
Figure~\ref{fig:droid_kinematic_swap} shows that substituting paired and mismatched skeleton kinematic means increases reconstruction MAE by 0.53$\%$ and 11.67$\%$, respectively.
SSIM exhibits the same ordering.
Zeroing the kinematic component increases MAE by 5.78$\%$, compared with 0.61$\%$ for the residual.
The greater degradation under mismatched substitution than kinematic removal indicates sensitivity to motion correspondence.
These results support compatible RGB and skeleton conditioning and a functional role for the kinematic component in reconstruction.

\subsection{Zero-Shot Human-to-Robot Motion Transfer}
\label{sec:skelam_human_transfer}

\begin{figure}[h]
    \centering
    \includegraphics[width=\linewidth]{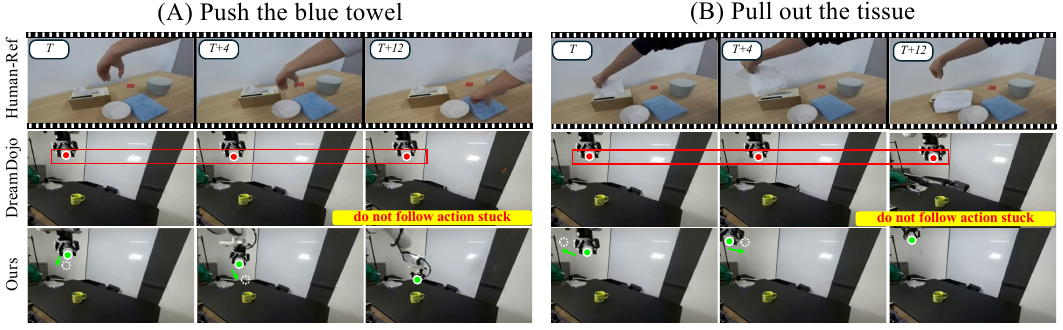}
    \caption{
        \textbf{Zero-shot human-to-robot motion transfer.}
        Human videos from UniSkill~\citep{kim2025uniskill} provide latent conditions for robot rollout generation.
        Markers indicate end-effector positions.
    }
    \label{fig:human_motion_transfer}
\end{figure}

To examine whether the learned motion representation transfers across embodiments, we use human demonstration videos from UniSkill~\citep{kim2025uniskill} as motion prompts.
The RGB encoder extracts latent actions from these videos to directly condition SkeLAM, with a robot observation specifying the initial scene.
All model parameters remain fixed during transfer.
We compare with the pretrained checkpoint of DreamDojo~\citep{gao2026dreamdojo}, which supports latent-action conditioning.
As shown in Figure~\ref{fig:human_motion_transfer}, SkeLAM generates distinct end-effector motions corresponding to the human demonstrations: a downward reach for the towel-pushing prompt and an upward movement for the tissue-pulling prompt.
DreamDojo exhibits limited end-effector displacement in both examples.
These results provide qualitative evidence that human-derived latent actions can guide robot motion despite differences in embodiment and scene appearance.

\begingroup
\setlength{\intextsep}{4pt plus 1pt minus 1pt}
\setlength{\textfloatsep}{4pt plus 1pt minus 1pt}
\setlength{\floatsep}{4pt plus 1pt minus 1pt}
\begin{table}[!htbp]
    \centering
    \caption{
        \textbf{Open-loop policy evaluation on RoboArena.}
        Seven policies are evaluated across 15 sessions (105 episodes).
        Episode-level metrics include balanced accuracy (BA, \%),
        success recall (\%), and Matthews correlation coefficient (MCC).
        Policy-level metrics include Spearman rank correlation ($\rho$),
        Pearson correlation ($r$), and success-rate MAE
        ($\Delta_{\mathrm{SR}}$).
    }
    \vspace{-2mm}
    \label{tab:policy_evaluation}

    \small
    \setlength{\tabcolsep}{3pt}
    \renewcommand{\arraystretch}{0.94}
    \begin{tabularx}{\linewidth}{@{}l*{6}{>{\raggedleft\arraybackslash}X}@{} }
        \toprule[1.02pt]
        & \multicolumn{3}{c}{Episode-level agreement}
        & \multicolumn{3}{c}{Policy-level agreement} \\
        \cmidrule(lr){2-4}\cmidrule(lr){5-7}
        Model
        & BA$\uparrow$
        & Recall$\uparrow$
        & MCC$\uparrow$
        & $\rho\uparrow$
        & $r\uparrow$
        & $\Delta_{\mathrm{SR}}\downarrow$ \\
        \midrule
        Cosmos-Predict2.5
        & 48.44 & 0.00 & -0.0525
        & -0.0756 & -0.1936 & 7.62 \\
        \rowcolor{orange!9} \textbf{SkeLAM}
        & 63.02 & 33.33 & 0.2483
        & 0.5347 & 0.6091 & 4.76 \\
        \bottomrule[1.02pt]
    \end{tabularx}
\end{table}

\vspace{-1mm}
\Needspace{12\baselineskip}
\subsection{World Model for Robot Policy Evaluation}
\label{sec:skelam_policy_evaluation}
\vspace{-1mm}

\begin{wrapfigure}{r}{0.3\columnwidth}
    \vspace{0pt}
    \centering
    \includegraphics[width=\linewidth]{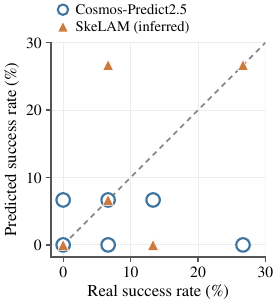}
    \captionsetup{skip=3pt}
    \caption{
        \small\textbf{{Policy-level agreement.}}    
        The dashed line indicates perfect agreement.
    }
    \label{fig:policy_agreement}
    \vspace{0pt}
\end{wrapfigure}
To assess whether generated rollouts preserve task outcomes and relative policy performance, we evaluate seven policies across 15 RoboArena sessions (105 episodes).
Both models receive the same initial observations and recorded actions.
Human annotators label task success in generated rollouts for comparison with real execution outcomes.
Table~\ref{tab:policy_evaluation} reports 33.33$\%$ success recall and 63.02$\%$ balanced accuracy for SkeLAM, compared with 0.00$\%$ and 48.44$\%$ for Cosmos-Predict2.5.
At the policy level, success-rate MAE decreases from 7.62 to 4.76 percentage points, with positive Spearman ($\rho$=0.5347) and Pearson ($r$=0.6091) correlations for SkeLAM.
Together, these metrics indicate improved agreement in both absolute policy success rates and relative policy ordering.
Figure~\ref{fig:policy_agreement} compares predicted and real success rates for individual policies, with proximity to the diagonal indicating closer agreement.
The results support agreement in both individual task outcomes and relative policy performance under open-loop action replay.
Nevertheless, the modest success recall indicates that reproducing successful task outcomes remains challenging.
Further evaluation details are provided in Appendix~\ref{app:policy_evaluation}.
\par
\clearpage
\endgroup

\vspace{-1mm}
\section{Conclusion}
\vspace{-1mm}

We presented SkeLAM, a skeleton-guided latent action model for controllable robot world modeling.
The kinematic component provides a motion-grounded target for robot action adaptation, while the residual retains complementary visual dynamics.
Cross-modal swap reconstruction and branch dropout support reuse of the learned kinematic pathway for action-conditioned generation.
Experiments on DROID show improved motion accuracy and visual fidelity, with representation analyses supporting kinematic specialization.
Qualitative results demonstrate zero-shot human-to-robot motion transfer, while policy evaluation yields success-rate estimates consistent with real robot executions.
These results support structuring latent actions around robot motion while retaining complementary information for generation.
Future work will extend the current recorded-action replay protocol to closed-loop policy evaluation.
A key question is how prediction errors affect subsequent policy decisions and task outcomes over longer interaction horizons.

\bibliography{iclr2027_conference}
\bibliographystyle{iclr2027_conference}

\clearpage
\appendix
\renewcommand{\eqref}[1]{(\ref{#1})}
\hypersetup{hidelinks}
\raggedbottom
\makeatletter
\setlength{\@fpsep}{3pt plus 1fil}
\makeatother
\captionsetup{font=small,skip=4pt}
\setlength{\floatsep}{5pt plus 1pt minus 1pt}
\setlength{\textfloatsep}{10pt plus 2pt minus 2pt}
\setlength{\intextsep}{6pt plus 2pt minus 2pt}
\renewcommand{\topfraction}{0.95}
\renewcommand{\bottomfraction}{0.9}
\renewcommand{\textfraction}{0.05}
\renewcommand{\floatpagefraction}{0.85}
\setcounter{topnumber}{6}
\setcounter{totalnumber}{6}
\section{Data and Preprocessing}
\label{app:skelam_video_data}

\paragraph{Data and splits.}
DROID and AgiBot World~\citep{bu2025agibot} are used in latent-action learning, world-model training, and action adaptation.
DROID contributes 60,000 training clips to Stage~1 and 45,358 clips to each of Stages~2A, 2B, 3A, and 3B.
RGB and skeleton observations are matched by episode, camera, and frame index; overlapping windows from an episode remain in the same partition.
Normalization statistics are fitted on training data only.
The DROID evaluation selects one clip from each of 400 trajectories: 100 trajectories per motion quartile, with additional camera stratification.
AgiBot evaluation uses 400 held-out clips and 800 arm trajectories.

\paragraph{Preprocessing.}
RGB and pre-rendered skeleton observations come from the corresponding OSCAR~\citep{wu2026oscar} subsets, following its kinematics, projection, and rasterization protocol.
Stage~1 letterboxes both modalities to $240\times320$ (height $\times$ width): resize by $\min(240/H_o,320/W_o)$ using OpenCV \texttt{INTER\_AREA}, round the resized dimensions, and center on a zero canvas, assigning odd padding pixels to the bottom or right.
Intensities are divided by 255.
The target skeleton mask selects pixels with an 8-bit RGB channel sum above 15, then applies one dilation with a $5\times5$ elliptical kernel.
It is used only in the skeleton loss.
Stage~2/3 uses a $4{:}3$ center crop resized to $480\times640$ for video generation; the frozen latent encoder keeps the Stage~1 resolution and uses the same source frame indices.

\paragraph{Temporal alignment and action interface.}
A DROID window samples 13 frames from 60-FPS video at indices $f_k=f_0+4k$, $k=0,\ldots,12$, giving 15 FPS and a 0.8-second span; consecutive window starts differ by 24 source frames.
The first frame conditions 12 future frames.
Each transition pairs $(o_k,o_{k+1})$ with an executed state-delta proxy
\[
 a_k=[p_{k+1}-p_k;\,\Delta\omega_k;\,g_{k+1}]\in\mathbb R^7,
\]
where translation is in the robot base frame, $\Delta\omega_k$ is the rotation vector of the relative $\mathrm{SO}(3)$ rotation, and $g_{k+1}$ is destination gripper openness.
Deltas are between consecutive sampled states, not relative to the clip start or joint-space commands.
Manifests retain source indices, timestamps, and transition-validity indicators.

\paragraph{Deterministic latent extraction.}
\label{app:latent_cache}
The frozen Stage~1 RGB encoder exports posterior means as two $12\times32$ arrays, $z_{\mathrm{kin}}$ and $z_{\mathrm{res}}$; the skeleton encoder is not used for world-model conditioning.
Each component is standardized elementwise as $\bar z=(z-m_z)\oslash\max(s_z,\epsilon)$ using training statistics.
The same kinematic statistics define adapter targets and world-model inputs; normalization is implicit in the main-text notation.
Offline caches retain episode, camera, frame, preprocessing, and checkpoint identifiers and receive no gradients.

\section{Model and Training Details}
\label{app:skelam_implementation}

\paragraph{Architecture and initialization.}
Both Transformer-VAE branches start from random weights and use $16\times16$ patches, width 1,024, 16 encoder and 16 decoder blocks, 16 attention heads, causal temporal attention, and zero Transformer dropout.
The RGB posterior has 32 kinematic and 32 residual dimensions; the skeleton posterior has 32 kinematic dimensions.
The world model starts from official Cosmos-Predict2.5 2B weights~\citep{ali2025world}, independently of the DreamDojo baseline.
The final Stage~1 teacher, video tokenizer, and text encoder remain frozen throughout Stage~2/3.

\begin{figure}[!htbp]
    \centering
    \begin{tikzpicture}[
        x=1cm,y=1cm,
        every node/.style={font=\scriptsize,align=center},
        stage/.style={draw=blue!45!black,fill=blue!4,rounded corners=2pt,
            text width=3.15cm,minimum height=0.78cm,inner sep=3pt},
        cache/.style={draw=black!50,dashed,fill=black!3,rounded corners=2pt,
            text width=3.15cm,minimum height=0.68cm,inner sep=3pt},
        source/.style={draw=black!45,fill=white,rounded corners=2pt,
            text width=3.15cm,minimum height=0.68cm,inner sep=3pt},
        weights/.style={-{Stealth[length=4pt]},semithick,draw=blue!45!black},
        data/.style={-{Stealth[length=4pt]},semithick,dashed,draw=black!60}
    ]
        \node[stage] (s1a) at (-4.2,0) {\textbf{Stage 1A} / 60K\\RGB + skeleton reconstruction};
        \node[stage] (s1b) at (0,0) {\textbf{Stage 1B} / 20K\\Grounding + alignment};
        \node[stage] (s1c) at (4.2,0) {\textbf{Stage 1C} / 5K\\Swap; then freeze teacher};

        \node[source] (cosmos) at (-4.2,-1.15) {Official Cosmos-Predict2.5 2B};
        \node[cache] (latents) at (0,-1.15) {Latent cache\\Frozen RGB teacher};
        \node[cache] (actions) at (4.2,-1.15) {Action cache\\Teacher latents + robot states};

        \node[stage] (s2a) at (-4.2,-2.3) {\textbf{Stage 2A} / 5K\\Conditioning / projector only};
        \node[stage] (s2b) at (0,-2.3) {\textbf{Stage 2B} / 20K\\Full DiT fine-tuning};
        \node[stage] (s3a) at (4.2,-2.3) {\textbf{Stage 3A} / 2K\\Action adapter fitting};
        \node[stage,fill=blue!10] (s3b) at (0,-3.5) {\textbf{Stage 3B} / 30K\\Full DiT +\\residual curriculum};

        \draw[weights] (s1a) -- (s1b);
        \draw[weights] (s1b) -- (s1c);
        \draw[data] (s1c.south) -- (4.2,-0.59) -| (latents.north);
        \draw[data] (latents) -- (actions);
        \draw[data] (latents.south) -- (0,-1.74) -| ([xshift=10pt]s2a.north);
        \draw[weights] (cosmos.south) -- (s2a.north);
        \draw[weights] (s2a) -- (s2b);
        \draw[data] (actions.south) -- (s3a.north);
        \draw[weights] (s2b.south) -- node[left,font=\scriptsize] {world model} (s3b.north);
        \draw[weights] (s3a.south) |- node[pos=0.7,above,font=\scriptsize] {adapter} (s3b.east);

        \draw[weights] (-5.75,-3.22) -- (-5.15,-3.22);
        \node[anchor=west] at (-5.05,-3.22) {Model weights};
        \draw[data] (-5.75,-3.68) -- (-5.15,-3.68);
        \node[anchor=west] at (-5.05,-3.68) {Cached data};
    \end{tikzpicture}
    \caption{\textbf{Training workflow.}
    Solid arrows transfer model weights; dashed arrows indicate cached data.
    Stage~3B starts from the Stage~2B world model and Stage~3A adapter and updates all DiT backbone weights.
    Cache construction does not add optimization stages.}
    \label{fig:app_training_workflow}
\end{figure}
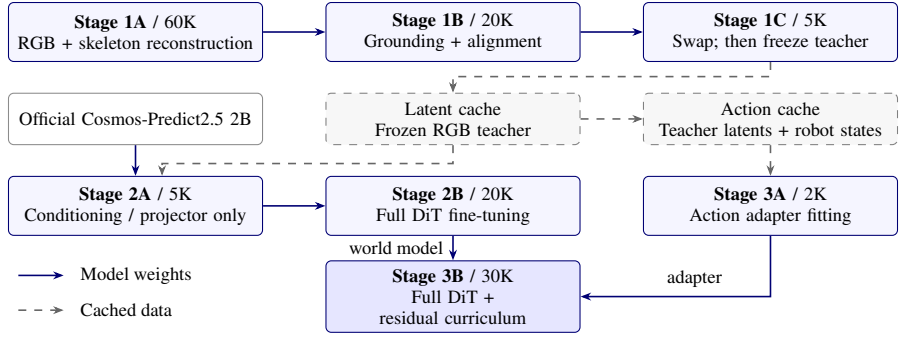

Figure~\ref{fig:app_training_workflow} summarizes the training stages, selected checkpoints, and dependencies.

\subsection{Latent-action training}
\label{app:stage1_training_details}

Stages~1A/B/C initialize RGB and skeleton reconstruction, add grounding and alignment, then add swap, respectively (Figure~\ref{fig:app_training_workflow}; $\mathrm{K}=1{,}000$).
The skeleton branch is frozen in Stage~1B/C.
RGB reconstruction uses MSE and the two KL terms in Eq.~\eqref{eq:rgb_variational_objective}, with $\beta=2\times10^{-4}$; swapping adds no KL term.
Grounding uses the RGB kinematic posterior mean through the frozen skeleton decoder; alignment uses paired, $\ell_2$-normalized means, with the skeleton target detached.
Swap reconstruction uses the frozen skeleton mean and the same residual sample as ordinary RGB reconstruction.

For target skeleton image $s$, foreground mask $m$, image prediction $\hat s$, and mask prediction $\hat m$, the grounding criterion is
\begin{align}
 \ell_{\mathrm{ske}}&=\ell_{\mathrm{fg}}+0.5\ell_{\mathrm{mask}}+0.05\ell_{\mathrm{bg}},\label{eq:app_skeleton_loss}\\
 \ell_{\mathrm{fg}}&=\frac{\sum_{p,c}m_p|\hat s_{pc}-s_{pc}|}{C_s\sum_p m_p+\epsilon},\qquad
 \ell_{\mathrm{bg}}=\frac{\sum_{p,c}(1-m_p)|\hat s_{pc}|}{C_s\sum_p(1-m_p)+\epsilon},\nonumber\\
 \ell_{\mathrm{mask}}&=\operatorname{BCE}(\hat m,m)+1-
 \frac{2\sum_p\hat m_pm_p+\epsilon}{\sum_p\hat m_p+\sum_pm_p+\epsilon},\nonumber
\end{align}
where $C_s$ is the number of color channels.
For batch representations $Z$, the regularizer is
\begin{align}
 V_{0.5}(Z)&=\frac{1}{d}\sum_{j=1}^{d}\max(0,0.5-\operatorname{Std}_b Z_{bj}),\nonumber\\
 \mathcal L_{\mathrm{var}}&=\frac{V_{0.5}(Z_{\mathrm{kin}})+V_{0.5}(Z_{\mathrm{res}})}{2}
 +\frac{0.1\|\operatorname{Cov}(Z_{\mathrm{kin}},Z_{\mathrm{res}})\|_F^2}{d_{\mathrm{kin}}d_{\mathrm{res}}}.
 \label{eq:app_variance_regularizer}
\end{align}
Cross-covariance uses centered RGB components and excludes the frozen skeleton latent; this discourages linear redundancy, without enforcing independence.
Stage~1B minimizes $\mathcal L_{\mathrm{rgb}}+\mathcal L_{\mathrm{ground}}+0.5\mathcal L_{\mathrm{align}}+0.1\mathcal L_{\mathrm{var}}$.
Stage~1C adds $\lambda_{\mathrm{swap}}\mathcal L_{\mathrm{swap}}$, with weight 0.5 for 500 updates and 1.0 thereafter; 20\% of updates instead use only $\mathcal L_{\mathrm{rgb}}$, including its KL terms.

Stages~1B/C use AdamW with $(\beta_1,\beta_2)=(0.9,0.999)$, $\epsilon=10^{-8}$, zero weight decay, gradient-norm clipping 0.3, and constant learning rates $5\times10^{-5}$ and $2\times10^{-5}$, respectively, without warmup.
The batch is 32 frame pairs on eight A100 80-GB GPUs (four pairs per GPU); Stage~1C uses bfloat16.
At phase transitions only model weights are restored.
The configured limits are 40K/10K updates, but the selected checkpoints are 20K/5K.

\subsection{World-model training and action adaptation}
\label{app:stage2_details}
\label{app:stage3_details}

Stage~2A performs 5K conditioning-projector updates, followed by 20K full-DiT and conditioning updates in Stage~2B.
Stage~3A fits only the action adapter for 2K updates with the world model fixed; its learning-rate cycle is configured for 8K updates.
Stage~3B starts from these Stage~2B/3A checkpoints and performs 30K action-conditioned generation updates.

\paragraph{Branch dropout.}
Standardized latents are masked before projection; the carrier contains two 32-dimensional vectors and two binary presence masks (66 values).
A dropped branch has zero latent and zero mask.
One mode is sampled per clip and shared over all 12 transitions.
For modes $(1,1),(0,0),(0,1),(1,0)$, probabilities are $(0.80,0.10,0.05,0.05)$ for global Stage~2 steps 0--999 and $(0.50,0.10,0.05,0.35)$ thereafter.
Stage~2B starts at global step 5,000.

\paragraph{Adapter and residual curriculum.}
The DROID adapter is a per-transition MLP, $7\rightarrow256\rightarrow256\rightarrow32$, with GELU between linear layers.
Training-standardized actions predict standardized RGB kinematic posterior means.
With valid-transition indicator $v_{ni}$, its regression loss is
\begin{equation}
 \mathcal L_{\mathrm{adapt}}=\frac{\sum_{n,i}v_{ni}\|A_\psi(a_{ni})-
 \operatorname{sg}[\mu^{\mathrm{rgb}}_{\mathrm{kin},ni}]\|_2^2}{32\sum_{n,i}v_{ni}+\epsilon}.
\end{equation}
In Stage~3B, kinematic conditions always come from actions with mask one.
The paired residual is retained with probability 0.5 for updates 0--1,999, decaying linearly to zero over updates 2,000--7,999, and zero thereafter.
One choice is shared across a clip; discarded residuals and their masks are zero.

\begin{table}[!htbp]
 \centering\small
 \caption{\textbf{World-model optimization.} Learning rates are module-specific.}
 \label{tab:app_world_training}
 \setlength{\tabcolsep}{4pt}\renewcommand{\arraystretch}{1.05}
 \begin{tabularx}{\linewidth}{@{}lLL@{}}
 \toprule
 Setting & Stage~2B & Stage~3B \\
 \midrule
 Trainable & Full DiT; both conditioning branches & Full DiT; adapter; $P_{\mathrm{kin}}$ \\
 Learning rates & DiT: $2.5\times10^{-5}$; conditioning: $5\times10^{-5}$ & DiT: $2.5\times10^{-5}$; adapter: $3\times10^{-5}$; $P_{\mathrm{kin}}$: $10^{-5}$ \\
 Residual projection & Trainable & Frozen \\
 \bottomrule
 \end{tabularx}
\end{table}
Both stages use AdamW, $(0.9,0.999)$, $\epsilon=10^{-8}$, weight decay $10^{-3}$, global gradient clipping 0.1, batch 32, bfloat16, eight-way FSDP, EMA, and 1K checkpoint intervals.
Stage~2B uses seed 0 and LambdaLinear scheduling: 100-step warmup, $f_{\mathrm{start}}=10^{-6}$, $f_{\max}=0.99$, $f_{\min}=0.4$, cycle 400K.
Normalization statistics remain fixed in Stage~3B.

\paragraph{DROID inference.}
One observed frame and 12 action transitions generate 12 future frames at $480\times640$, always with zero residual and mask.
We use \texttt{FlowUniPCMultistepScheduler}, order 2, solver \texttt{bh2}, 35 steps, shift 5, and a non-Karras schedule.
Guidance is $f_{\mathrm{cond}}+g(f_{\mathrm{cond}}-f_{\mathrm{uncond}})$ with $g=0$ (conditional prediction); text is the empty string encoded online, with negative prompts disabled.
Results use regular, not EMA, weights at Stage~3B step 30K.
Each clip is generated once with seed $20260823+p$ for zero-based manifest position $p$; eight ranks partition clips without padding or duplication.
All 400 generations completed.

\section{Evaluation Protocols}
\label{app:skelam_protocol}

\subsection{Video and motion metrics}
\label{app:skelam_video_metrics}

\paragraph{Video fidelity.}
Canonical clips have 13 RGB frames in $[0,1]$ at $480\times640$.
Video metrics include the initial conditioning frame; motion metrics use the 12 future frames.
MSE and MAE average errors over all channels, pixels, and frames per clip, then over valid clips.
PSNR is computed from each clip's MSE $e_n$ before averaging: $N_v^{-1}\sum_n10\log_{10}(1/e_n)$ dB.
SSIM uses \texttt{skimage.metrics.structural\_similarity}, \texttt{data\_range=1.0}, \texttt{channel\_axis=-1}, and the default $7\times7$ uniform window.
LPIPS uses TorchMetrics AlexNet with \texttt{normalize=True}.
Both are averaged over frames and then clips.
Temporal LPIPS is the mean over clips of
\begin{equation}
 \mathrm{tLPIPS}_n=\frac{1}{12}\sum_{t=1}^{12}\left|
 D_{\mathrm{LPIPS}}(\hat I_{n,t},\hat I_{n,t-1})-
 D_{\mathrm{LPIPS}}(I_{n,t},I_{n,t-1})\right|.
\end{equation}
It measures agreement in adjacent-frame perceptual change, not end-effector motion.

\paragraph{Validity and displacement.}
A valid arm trajectory requires in-image reference positions and complete prediction tracking at every evaluated frame, with finite coordinates throughout.
Invalid trajectories are excluded from scores but remain in the coverage denominator (400 for DROID, 800 for AgiBot).
Thus methods can have different valid subsets; generated-video and valid-video counts are reported separately.
For valid trajectories, let $\Delta p_{nt}=p_{nt}-p_{n0}$ and $\Delta\hat p_{nt}=\hat p_{nt}-\hat p_{n0}$.
Then $\mathrm{ADE}=(12N_m)^{-1}\sum_{n=1}^{N_m}\sum_{t=1}^{12}\|\Delta\hat p_{nt}-\Delta p_{nt}\|_2$ pixels.

\paragraph{Direction and magnitude.}
Direction uses initial-to-final displacement: the sign of the horizontal component determines left/right when $|\Delta x|\geq|\Delta y|$; otherwise the vertical sign determines up/down (positive image $y$ is down).
Displacements below 0.005 image diagonals (4 pixels) are stationary.
Four-class direction macro-F1 includes only trajectories for which both reference and prediction are nonstationary, a potentially smaller subset than ADE.
Magnitude is $m_n=\|\Delta p_{n,12}\|_2/800$; magnitude MAE is $N_m^{-1}\sum_n|\hat m_n-m_n|$, and $r_{\mathrm{mag}}$ and $\rho_{\mathrm{mag}}$ are Pearson and Spearman correlations over valid trajectories.
Undefined correlations are unavailable, not zero.

\subsection{Baselines, probes, and interventions}

Table~\ref{tab:main_results} lists control interfaces; text-only TesserAct (RGB-LoRA) and GE-Base are visual references.
Linear probes use frozen encoders and $\widehat{\Delta p}=Wz+b$: displacement MAE/$R^2$ and magnitude macro-F1.
Individual latents are 32D, their concatenation 64D; the skeleton teacher uses privileged input.

For interventions, deterministic means $k,r,s$ denote RGB kinematic, RGB residual, and skeleton kinematic latents.
Normal, correct/wrong swap, and zero-kinematic/residual inputs are $[k;r]$, $[s;r]$, $[s';r]$, $[0;r]$, and $[k;0]$; $s'$ comes from another episode.
The decoder, current frame, and target are fixed.
Figure~\ref{fig:droid_kinematic_swap} reports $100(\bar e_j-\bar e_0)/\bar e_0$ for MAE and $10^3(\overline{\mathrm{SSIM}}_0-\overline{\mathrm{SSIM}}_j)$.
Raw decoder-latent zeroing differs from world-model dropout of standardized latents and presence masks.
These tests support motion specialization without establishing complete disentanglement.

\subsection{Human transfer and policy evaluation}
\label{app:human_transfer}
\label{app:policy_evaluation}

\paragraph{Human-to-robot transfer.}
UniSkill~\citep{kim2025uniskill} videos provide RGB-encoder latent sequences directly to the world model, bypassing the action adapter; a robot observation specifies the initial scene.
All weights remain fixed, and DreamDojo uses its latent-conditioned checkpoint.
The two examples assess visible motion correspondence, not task completion or reliable transfer across arbitrary viewpoints and embodiments.

\paragraph{Open-loop policy evaluation.}
Seven RoboArena policies are evaluated in 15 sessions (105 paired episodes), with nine real successes and 96 failures.
Policy identifiers are \texttt{pi0\_fast}, \texttt{paligemma\_fast}, \texttt{paligemma\_fast\_specialist}, \texttt{paligemma\_vq}, \texttt{pi0}, \texttt{paligemma\_binning}, and \texttt{paligemma\_diffusion}, each with suffix \texttt{\_droid}.
Models receive the same initial observations and recorded actions; human annotators assign binary task-success labels to generated rollouts using the real-execution criterion.
Actions are replayed without policy feedback from generated observations.
Episode-level metrics pool all pairs, treating success as positive: success recall, balanced accuracy (mean sensitivity and specificity), and MCC.
For policy $p$, real and generated success rates $r_p,\hat r_p$ average the 15 session labels.
Pearson and Spearman correlations use seven rate pairs, with average ranks for ties; success-rate MAE is $\Delta_{\mathrm{SR}}=(100/7)\sum_p|r_p-\hat r_p|$ percentage points.
The seven policies and nine real successes limit the ranking evidence.

\section{Supplementary Results and Ablations}
\label{app:additional_experiments}

\paragraph{Additional metrics and coverage.}
Table~\ref{tab:app_droid_extra} complements Table~\ref{tab:main_results} with DROID metrics and trajectory coverage.
OSCAR has lower tLPIPS than SkeLAM (0.0174 vs.\ 0.0222), while SkeLAM has lower ADE and LPIPS.
Tables~\ref{tab:app_agibot_visual} and~\ref{tab:app_agibot_motion} report AgiBot results: SkeLAM improves SSIM over DreamDojo but has worse LPIPS and ADE, so DROID gains do not extend uniformly to AgiBot.
Table~\ref{tab:app_adaptation_budget} compares adaptation budgets.

\paragraph{Component ablations.}
In Table~\ref{tab:ablations}, removing skeleton supervision omits grounding, alignment, and swap, retaining RGB reconstruction; removing swap retains grounding and alignment with $\lambda_{\mathrm{swap}}=0$.
Removing the residual uses zero residual from Stage~2 onward.
Removing action adaptation skips adapter regression and jointly trains a randomly initialized adapter and world model for action-conditioned generation.
The latter two retain the full model's action interface and inference settings, including zero residual.
These change training; decoder interventions keep weights fixed.

\begin{table}[!htbp]
\centering\small
\setlength{\tabcolsep}{2.5pt}\renewcommand{\arraystretch}{1.00}
\caption{\textbf{DROID: additional metrics and coverage.} All methods generate 400 valid videos. Primary metrics appear in Table~\ref{tab:main_results}; $N_m$ counts valid arm trajectories out of 400.}
\label{tab:app_droid_extra}
\begin{tabular}{@{}lrrrr@{}}
\toprule
Model & $N_m$ & MAE$\downarrow$ & tLPIPS$\downarrow$ & $\rho_{\mathrm{mag}}\uparrow$ \\
\midrule
TesserAct & 315/400 & 0.0928 & 0.0993 & 0.0431 \\
GE-Base & 330/400 & 0.1587 & 0.0549 & 0.0340 \\
IRASim & 328/400 & 0.1917 & 0.0912 & 0.1318 \\
Ctrl-World & 326/400 & 0.0859 & 0.0220 & 0.4980 \\
Cosmos-Predict2.5 & 326/400 & 0.1873 & 0.0596 & 0.0351 \\
Cosmos3 & 328/400 & 0.0753 & 0.0570 & 0.2080 \\
EnerVerse-AC & 327/400 & 0.1208 & 0.0584 & -0.0448 \\
GE-Sim & 328/400 & 0.1578 & 0.0534 & 0.0346 \\
OSCAR & 325/400 & 0.1531 & 0.0174 & 0.5030 \\
DreamDojo & 325/400 & 0.0532 & 0.0366 & 0.3032 \\
CD-LAM & 325/400 & 0.0571 & 0.0352 & 0.2496 \\
SkeLAM & 326/400 & 0.0471 & 0.0222 & 0.5283 \\
\bottomrule
\end{tabular}
\end{table}

\begin{table}[!htbp]
\centering\small
\setlength{\tabcolsep}{2.5pt}\renewcommand{\arraystretch}{1.00}
\caption{\textbf{AgiBot World: video fidelity.} All methods generate 400 videos; $N_v$ is the valid-video count. MSE/MAE use normalized RGB; PSNR is in dB.}
\label{tab:app_agibot_visual}
\begin{tabular}{@{}lrrrrrrr@{}}
\toprule
Model & $N_v$ & MSE$\downarrow$ & MAE$\downarrow$ & PSNR$\uparrow$ & SSIM$\uparrow$ & LPIPS$\downarrow$ & tLPIPS$\downarrow$ \\
\midrule
TesserAct & 376 & 0.0223 & 0.0684 & 17.19 & 0.7781 & 0.2043 & 0.0673 \\
GE-Base & 400 & 0.0077 & 0.0392 & 22.15 & 0.7732 & 0.3004 & 0.0268 \\
IRASim & 400 & 0.1889 & 0.3568 & 7.25 & 0.1489 & 0.8636 & 0.5998 \\
Ctrl-World & 376 & 0.0364 & 0.1146 & 14.69 & 0.5471 & 0.5313 & 0.0243 \\
Cosmos-Predict2.5 & 400 & 0.0576 & 0.1627 & 12.67 & 0.4610 & 0.5950 & 0.0408 \\
Cosmos3 & 400 & 0.0151 & 0.0523 & 19.74 & 0.7982 & 0.1740 & 0.0315 \\
EnerVerse-AC & 376 & 0.0355 & 0.1113 & 14.81 & 0.5425 & 0.4484 & 0.0185 \\
GE-Sim & 376 & 0.0073 & 0.0383 & 22.66 & 0.8459 & 0.1506 & 0.0260 \\
OSCAR & 376 & 0.0058 & 0.0329 & 23.44 & 0.8399 & 0.0966 & 0.0126 \\
DreamDojo & 400 & 0.0067 & 0.0341 & 22.79 & 0.8330 & 0.0889 & 0.0134 \\
CD-LAM & 400 & 0.0092 & 0.0421 & 20.97 & 0.7994 & 0.1172 & 0.0113 \\
SkeLAM & 400 & 0.0088 & 0.0344 & 21.79 & 0.8479 & 0.1049 & 0.0181 \\
\bottomrule
\end{tabular}
\end{table}
\begin{table}[!htbp]
\centering\small
\setlength{\tabcolsep}{2.5pt}\renewcommand{\arraystretch}{1.00}
\caption{\textbf{AgiBot World: motion and coverage.} $N_m$ counts valid arm trajectories out of 800. ADE is in pixels and magnitude error is normalized by the image diagonal.}
\label{tab:app_agibot_motion}
\begin{tabular}{@{}lrrrrrr@{}}
\toprule
Model & $N_m$ & ADE$\downarrow$ & $\mathrm{F1}_{\mathrm{dir}}\uparrow$ & $L_1^{\mathrm{mag}}\downarrow$ & $r_{\mathrm{mag}}\uparrow$ & $\rho_{\mathrm{mag}}\uparrow$ \\
\midrule
TesserAct & 684/800 & 45.74 & 0.2348 & 0.0850 & -0.0489 & 0.0548 \\
GE-Base & 754/800 & 16.71 & 0.4830 & 0.0336 & 0.1224 & 0.5104 \\
IRASim & 750/800 & 78.81 & 0.2628 & 0.1182 & -0.0335 & 0.0027 \\
Ctrl-World & 707/800 & 18.52 & 0.2652 & 0.0382 & 0.1127 & 0.0317 \\
Cosmos-Predict2.5 & 754/800 & 26.61 & 0.2802 & 0.0481 & 0.0491 & 0.0816 \\
Cosmos3 & 752/800 & 28.61 & 0.2853 & 0.0477 & 0.0972 & 0.4452 \\
EnerVerse-AC & 709/800 & 18.27 & 0.2276 & 0.0376 & 0.0442 & 0.1576 \\
GE-Sim & 707/800 & 17.22 & 0.2529 & 0.0359 & 0.0331 & 0.3021 \\
OSCAR & 708/800 & 17.32 & 0.5557 & 0.0309 & 0.1946 & 0.6831 \\
DreamDojo & 754/800 & 19.55 & 0.5720 & 0.0338 & 0.2056 & 0.6299 \\
CD-LAM & 750/800 & 22.73 & 0.5273 & 0.0394 & 0.1583 & 0.5983 \\
SkeLAM & 755/800 & 21.10 & 0.2824 & 0.0366 & 0.1188 & 0.5397 \\
\bottomrule
\end{tabular}
\end{table}

\begin{table}[!htbp]
\centering\small
\setlength{\tabcolsep}{2.5pt}\renewcommand{\arraystretch}{1.00}
\caption{\textbf{Stage~3B adaptation budgets.} Each row uses 400 videos; $N_m$ lists valid/total arm trajectories. Budgets exclude pretraining and the common 2K adapter fit; inference uses zero residual. DROID LPIPS decreases with updates, but motion metrics on both datasets are nonmonotonic.}
\label{tab:app_adaptation_budget}
\begin{tabular}{@{}llrrrrrrr@{}}
\toprule
Data & Steps & $N_m$ & ADE$\downarrow$ & SSIM$\uparrow$ & LPIPS$\downarrow$ & tLPIPS$\downarrow$ & $\mathrm{F1}_{\mathrm{dir}}\uparrow$ & $r_{\mathrm{mag}}\uparrow$ \\
\midrule
DROID & 5K & 325/400 & 45.25 & 0.8299 & 0.1374 & 0.0241 & 0.5257 & 0.4930 \\
 & 10K & 326/400 & 45.37 & 0.8326 & 0.1337 & 0.0232 & 0.5248 & 0.4986 \\
 & 15K & 327/400 & 42.48 & 0.8345 & 0.1314 & 0.0228 & 0.5439 & 0.5408 \\
 & 30K & 326/400 & 41.15 & 0.8340 & 0.1310 & 0.0222 & 0.5465 & 0.5277 \\
AgiBot & 5K & 756/800 & 20.79 & 0.8452 & 0.1073 & 0.0185 & 0.2968 & 0.1742 \\
 & 10K & 754/800 & 21.29 & 0.8477 & 0.1056 & 0.0182 & 0.2888 & 0.1332 \\
 & 15K & 755/800 & 21.14 & 0.8471 & 0.1057 & 0.0182 & 0.2752 & 0.1639 \\
 & 30K & 755/800 & 21.10 & 0.8479 & 0.1049 & 0.0181 & 0.2824 & 0.1188 \\
\bottomrule
\end{tabular}
\end{table}

\clearpage
\section{Additional Qualitative Comparisons}
\label{app:qualitative_comparisons}

\begin{figure}[!htbp]
    \centering
    \includegraphics[width=\linewidth]{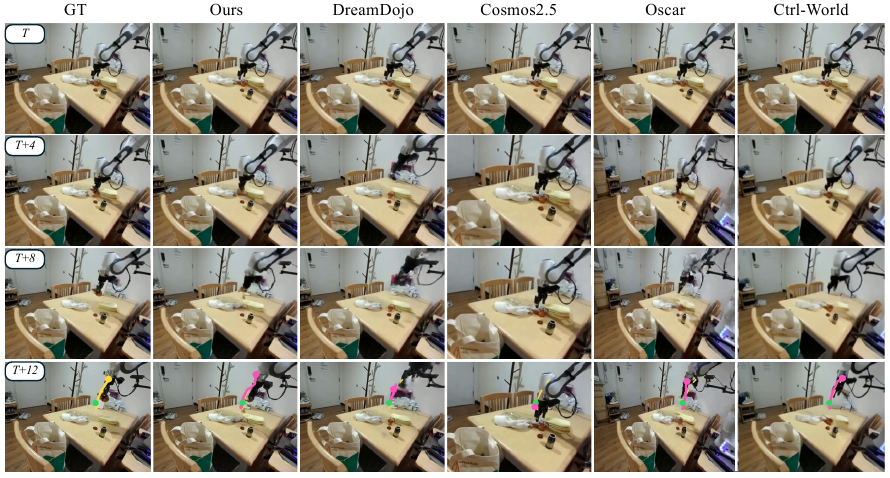}
    \caption{\textbf{Tabletop scene.}
    Columns: ground truth, SkeLAM, DreamDojo, Cosmos-Predict2.5, OSCAR, and Ctrl-World.
    Rows: $T$, $T+4$, $T+8$, and $T+12$.
    Colored markers and arrows highlight end-effector motion in the final row.}
    \label{fig:app_qualitative_tabletop}
    \par\medskip
    \includegraphics[width=\linewidth]{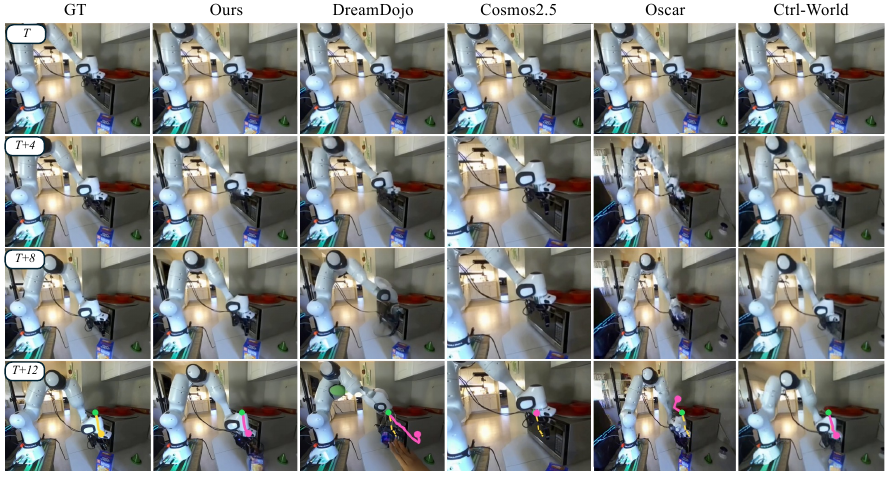}
    \caption{\textbf{Kitchen scene.}
    Models and time steps follow Figure~\ref{fig:app_qualitative_tabletop}; markers and arrows highlight end-effector motion in the final row.}
    \label{fig:app_qualitative_countertop}
\end{figure}

Figures~\ref{fig:app_qualitative_tabletop} and~\ref{fig:app_qualitative_countertop} supplement Figure~\ref{fig:qualitative_results} with two DROID scenes.
Both grids compare recorded observations and generated rollouts at four time steps, showing end-effector motion, arm geometry, and scene appearance.

\paragraph{AI use statement.}
Generative AI assisted language polishing and readability; all AI-assisted content was reviewed by authors.

\end{document}